\documentclass[10pt,twocolumn,letterpaper]{article}

\usepackage{cvpr}
\usepackage{times}
\usepackage{epsfig}
\usepackage{graphicx}
\usepackage{amsmath}
\usepackage{amssymb}
\usepackage{authblk}

\usepackage[breaklinks=true,bookmarks=false]{hyperref}

\cvprfinalcopy 

\def\cvprPaperID{****} 

\begin{document}

\title{Rethinking the Heatmap Regression for Bottom-up Human Pose Estimation }
\author[1,2,3,4]{\thanks{This work is done when Zhengxiong is an intern at MEGVII Research.} Zhengxiong Luo}
\author[1]{Zhicheng Wang}
\author[3,4]{Yan Huang}
\author[3,4]{Liang Wang}
\author[3]{Tieniu Tan}
\author[1]{Erjin Zhou}

\affil[1]{
	Megvii Inc \quad $^2$University of Chinese Academy of Sciences (UCAS)
}
\affil[3]{
	Center for Research on Intelligent Perception and Computing (CRIPAC)\authorcr
	National Laboratory of Pattern Recognition (NLPR)
}
\affil[4]{
	Institute of Automation, Chinese Academy of Sciences (CASIA) \authorcr
	\normalsize{zhengxiong.luo@cripac.ia.ac.cn \quad \{wangzhicheng, zej\}@megvii.com \quad \{yhuang, wangliang, tnt\}@nlpr.ia.ac.cn}
}

\maketitle

	\begin{abstract}
	
	Heatmap regression has become the most prevalent choice for nowadays human pose estimation methods. The ground-truth heatmaps are usually constructed via covering all skeletal keypoints by 2D gaussian kernels. The standard deviations of these kernels are fixed. However, for bottom-up methods, which need to handle a large variance of human scales and labeling ambiguities, the current practice seems unreasonable. To better cope with these problems, we propose the scale-adaptive heatmap regression (SAHR) method, which can adaptively adjust the standard deviation for each keypoint. In this way, SAHR is more tolerant of various human scales and labeling ambiguities. However, SAHR may aggravate the imbalance between fore-background samples, which potentially hurts the improvement of SAHR. Thus, we further introduce the weight-adaptive heatmap regression (WAHR) to help balance the fore-background samples. Extensive experiments show that SAHR together with WAHR largely improves the accuracy of bottom-up human pose estimation. As a result, we finally outperform the state-of-the-art model by $+1.5AP$ and achieve  $72.0 AP$ on COCO test-dev2017, which is comparable with the performances of most top-down methods.  Source codes are available at \url{https://github.com/greatlog/SWAHR-HumanPose}.
	
\end{abstract}

\section{Introduction}

Multi-person human pose estimation (HPE) aims to locate skeletal keypoints of all persons in a given RGB image. It has been widely applied in human activity recognition, human computer interaction, animation~\etc. Current human pose estimation methods fall into two categories: \textit{top-down} and \textit{bottom-up}. In top-down methods, all persons are firstly cropped out by a human detector and then resized to the same size before they are input to the keypoints detector. Oppositely, bottom-up methods directly detect keypoints of all persons simultaneously. It is more light-weight fast but suffers from various human scales.

\begin{figure}
	\centering
	\includegraphics[width=\linewidth]{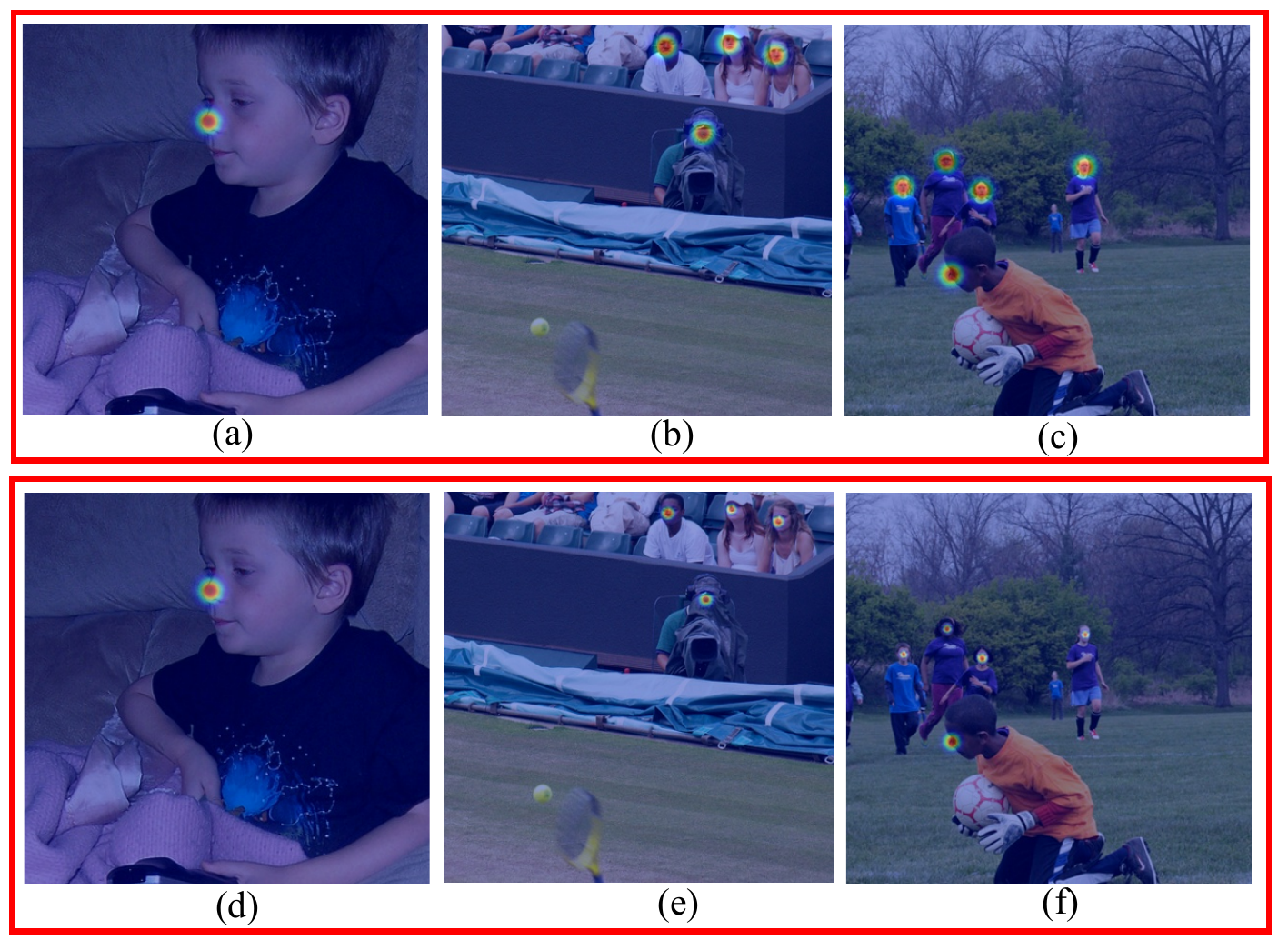}
	\caption{Top row: the noses of different persons are covered by gaussian kernels with the same standard deviation. Bottom row: the standard deviations for keypoints of different persons are adaptively adjusted in SAHR.}\label{nose_heatmap}
\end{figure} 

Heatmap regression is widely used in both top-down and bottom-up HPE methods. The ground-truth heatmaps are constructed by putting 2D Gaussian kernels on all keypoints. They are used to supervise the predicted heatmaps via L2 loss. This method is easy to be implemented and has much higher accuracy than traditional coordinate regression~\cite{deeppose,integralpose,pose_proposal}. However, in current practice, different keypoints are covered by gaussian kernels with the same standard deviation~\cite{cpm,cpn,stacked}, which means that different keypoints are supervised by the same constructed heatmaps.

We argue that this is unreasonable in two aspects. Firstly, keypoints of different scales are semantically discriminative in regions of different spatial sizes. It may cause confusion to put the same gaussian kernel on all keypoints. As shown in the top row of Figure~\ref{nose_heatmap}, the noses of different persons are covered by gaussian kernels with the same deviation ($\sigma=2$). In (a), the covered region is restricted on the top of the nose. But in (b), the Gaussian kernel could cover the face, and in (c), the whole head is even covered. The various covered regions for the same keypoint may cause semantic confusion. Secondly, even humans could not label the keypoints with pixel-wise accuracy, and the ground-truth coordinates may have inherent ambiguities~\cite{BoundingBR,gaussianyolo}. Thus the keypoints could be treated as distributions (instead of discrete points) centered around the labeled coordinates. Their standard deviations represent their uncertainties and should be proportion to the labeling ambiguities. However, current practice keeps the standard deviations fixed. It seems to have ignored the influence of various labeling ambiguities of different keypoints.

From the above discussion, the standard deviation for each keypoint should be related to its scale and uncertainty. A straightforward way to solve these issues is manually labeling different keypoints with different standard deviations. However, this work is extremely labor-intensive and time-consuming. Besides, it is difficult to define customized standard deviations for different keypoints. Towards this problem, we propose the scale-adaptive heatmap regression (SAHR), which can adaptively adjust the standard deviation for each keypoint by itself. 

Specifically, we firstly cover all keypoints by Gaussian kernels of the same base standard deviation $\sigma_0$. We add a new branch to predict \textit{scale maps} $\mathbf{s}$, which are of the same shape as ground-truth heatmaps. Then we modify the original standard deviation for each keypoint to $\sigma_0\cdot\mathbf{s}$ by a point-wise operation. Thus to some extent, $\mathbf{s}$ represents the scales and uncertainties of corresponding keypoints. In this way, the suitable standard deviations for different keypoints could be adaptively learned, and thus SAHR may be more tolerant of various human scales and labeling ambiguities. However, as shown in the bottom row of Figure~\ref{nose_heatmap}, SAHR may aggravate the imbalance between fore-background samples, which potentially restricts the improvements of SAHR~\cite{focalloss,simplepose}. Motivated by focal loss for classification~\cite{focalloss}, we further introduce the weight-adaptive heatmap regression (WAHR), which can automatically down-weight the loss of relatively easier samples, and focus more on relatively harder samples. Experiments show that the improvements brought by SAHR can be further advanced by WAHR.

Our contributions can be summarized as four points:
\begin{itemize}
	\item [1.] To the best of our knowledge, this is the first paper that focuses on the problems in heatmap regression when tackling large variance of human scales and labeling ambiguities. We attempt to alleviate these problems by scale and uncertainty prediction.
	\item [2.] We propose a scale-adaptive heatmap regression (SAHR), which can adaptively adjust the standard deviation of the Gaussian kernel for each keypoint, enabling the model to be more tolerant of various human scales and labeling ambiguities. 
	\item[3.] We propose a weight-adaptive heatmap regression (WAHR) to alleviate the severe imbalance between foreground and background samples. It could automatically focus more on relatively harder examples and fully exploit the superiority of SAHR.
	\item[4.] Our model outperforms the state-of-the-art model by $1.5AP$ and achieves $72.0 AP$ on COCO test-dev2017, which is comparable with the performances of most top-down methods.
\end{itemize}

\section{Related Works}

\subsection{Bottom-up Human Pose Estimation}

Bottom-up HPE methods firstly detect all identity-free keypoints and then group them into individual persons. Compared with recent top-down HPE methods~\cite{cpn,simplebaseline,hrnet,rsn}, bottom-up methods are usually inferior on accuracy. However, since they do not rely on human detectors and could decouple the runtime with the number of persons, bottom-up methods may have more potential superiority on speed~\cite{openpose}. But on the other hand, bottom-up methods have to tackle the grouping problem and large variance of human scales. 

Recent works about bottom-up HPE mostly focus on developing better grouping methods~\cite{ae,personlab,simplepose,pipaf,oap}.  In~\cite{pipaf}, a Part Intensity Field (PIF) and a Part Association Field (PAF) are used to localize and associate body parts. In~\cite{simplepose}, the body parts are learned in the same way as keypoints by heatmaps. And in~\cite{oap}, keypoints are grouped according to their offsets from corresponding center points. In this paper, we use associative embedding proposed in~\cite{ae}, which simple yet proved to be effective for points grouping~\cite{pushpullloss1,pushpullloss2,cornet}. Although the grouping method has been advanced a lot, few works are done about the various human scales. In this paper, we mainly focus on the problems in bottom-up HPE when tackling large variance of human scales.

\subsection{Heatmap Regression}

Heatmap regression is widely used for semantic landmarks localization, such as keypoints of human faces~\cite{bulat2017far}, hands~\cite{multiviewhand}, bodies~\cite{cpm,stacked} and household objects~\cite{robot}. The ground-truth heatmaps are constructed by putting 2D Gaussian kernels on the labeled points. The pixel values on the heatmaps are usually treated as the probabilities of corresponding pixels being the keypoints. This method is easy to be implemented and could potentially attain pixel-wise accuracy. Thus heatmap regression has become the dominant method for HPE. However, current methods typically cover all keypoints by Gaussian kernels with the same standard deviations. It may work well for top-down methods, in which all persons are resized to the same size. But in bottom-up methods, in which persons are of various scales, it seems to be more desirable to adjust the standard deviation for each keypoint according to the scale of the corresponding person.

\begin{figure*}[t]
	\centering
	\includegraphics[width=\linewidth]{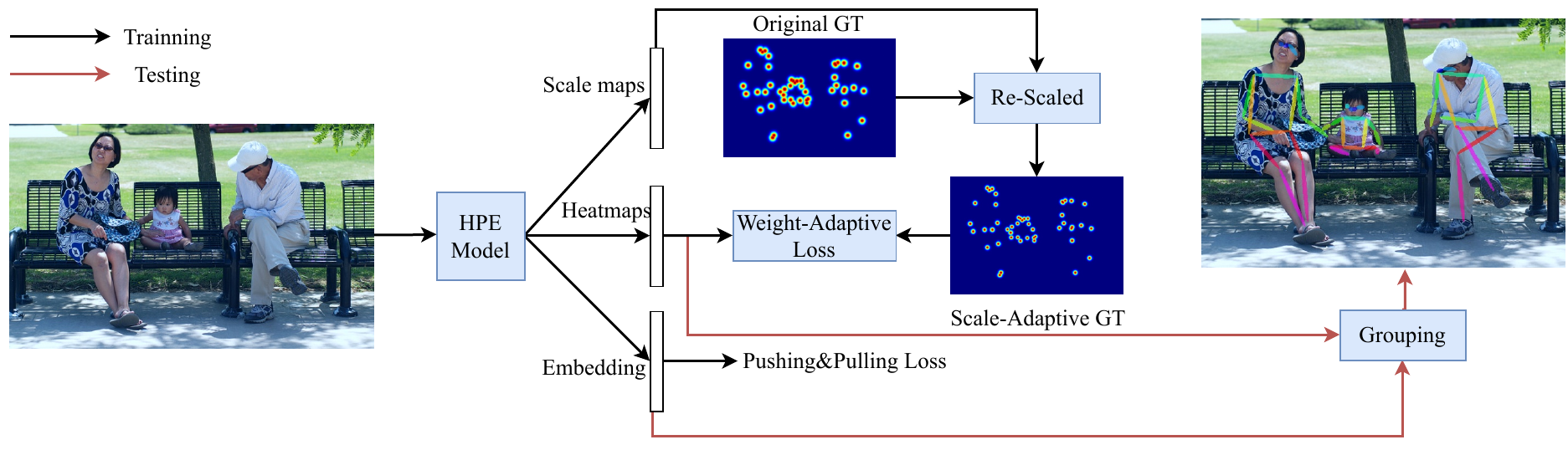}
	\caption{During training, the ground-truth heatmaps are firstly scaled according to predicted scale maps and then are used to supervise the whole model via weight-adaptive loss. During testing, the predicted heatmaps  and associative embeddings are used for grouping of individual persons.}\label{SWAHR_loss}
\end{figure*} 

\subsection{Uncertainty Prediction}

Recently, uncertainty prediction has become an important method for many tasks~\cite{cyz,Isobe2017DeepCE,Gal2016DropoutAA,Shi2019ProbabilisticFE}. As there are usually inevitable labeling ambiguities in the training datasets~\cite{Shi2019ProbabilisticFE}, it is better to explicitly model the uncertainty for predictions. In~\cite{BoundingBR}, He \etal treat the positions of ground-truth bounding boxes as Gaussian distributions around the labeled coordinates, and use KL loss~\cite{klloss} to supervise the model. In~\cite{gaussianyolo}, a similar idea is adopted to predict the coordinates and objecting scores of bounding boxes. For HPE, inherent ambiguities may also exist in ground-truth keypoint, such as inaccurate labeling, occlusion, or ambiguous cases. Original heatmap regression covers keypoints by Gaussian kernels while keeping standard deviations fixed. In that case, the ambiguities of different keypoints are assumed to be the same. This implicit assumption may be too strong and potentially hurt the performance. In this paper, the scale-adaptive heatmap regression alleviates this problem by introducing scale maps to adaptively modify the standard deviation for each keypoint.

\section{Proposed Method}

\subsection{Formulation}
Suppose $C^p_k = \{x^p_k, y^p_k\}$ denotes the coordinate of the $k^{th}$ keypoint of the $p^{th}$ person, and $\mathbf{h}^p$ denotes its corresponding ground-truth heatmap, then the covered region for $C^p_k$ is written as
\begin{equation}
	\begin{aligned}
		&\mathbf{h}^p_{k, i, j}= e^{{-((i-x^p_k)^2 + (j - y^p_k)^2)}/{2 \sigma^2}}& \\ 
		&s.t. \quad \|i - x^p_k\|_1 \leq 3\sigma \quad \|j - y^p_k\|_1 \leq 3\sigma, &
	\end{aligned}
\end{equation}
where $\sigma$ denotes the standard deviation, and $\{k, i, j\}$ indicates the position of pixel on  $\mathbf{h}^p$. For $\|i - x^p_k\| \textgreater 3\sigma$ or $\|j - y^p_k\| \textgreater 3\sigma$, we have $\mathbf{h}^p_{k, i, j}=0$. If the number of persons is $N$, then the overall ground-truth heatmaps are 
\begin{equation}
	\mathbf{H^{\sigma}} = \max\{\mathbf{h}^1, \mathbf{h}^2, \dots , \mathbf{h}^N\},
\end{equation}
where $\max$ is pixel-wisely operated. 

Suppose the predicted heatmaps are $\mathbf{P}$, then the regression loss is
\begin{equation}
	\mathcal{L}_{regressoin} = \| \mathbf{P} - \mathbf{H}^{\sigma} \|^2_2.
\end{equation}

\subsection{Scale-Adaptive Heatmap Regression}
In previous methods, the standard deviation $\sigma$ is fixed as $\sigma_0$ for all keypoints, in which case the ground-truth heatmaps are denoted as $\mathbf{H^{\sigma_0}}$. However, keypoints of different scales have semantically discriminative regions, thus they are expected to be covered by Gaussian kernels with different deviations. Since it is hard to manually label each keypoint, we hope that the model could learn to adjust $\sigma$ by itself. 

We add a new branch to predict the \textit{scale maps} $\mathbf{s}$,  which are of the shape with ground-truth heatmaps. For keypoint $C^p_k = \{x^p_k, y^p_k\}$, we modify the standard deviation to $\sigma_0\cdot\mathbf{s}_{k, x^p_k, y^p_k}$. then the covered region for $C^p_k$ becomes
\begin{equation}
	\begin{aligned}
		&\mathbf{h}^p_{k, i, j}= e^{{-((i-x^p_k)^2 + (j - y^p_k)^2)}/{2 (\sigma_0\cdot\mathbf{s}_{k, x^p_k, y^p_k})^2}}& \\ 
		&s.t. \quad \|i - x^p_k\|_1 \leq 3\sigma \quad \|j - y^p_k\|_1 \leq 3\sigma.&
	\end{aligned}
\end{equation}
Since the covered region is relatively small, we may have $\mathbf{s}_{k, x^p_k, y^p_k}\approx\mathbf{s}_{k, i, j}$ for  $\|i - x^p_k\|_1 \leq 3\sigma$ and  $\|j - y^p_k\|_1 \leq 3\sigma$. Thus, for simplicity, the modification can be written as an element-wise operation:
\begin{equation}
	\begin{aligned}
		&\mathbf{h}^p_{k, i, j}= e^{{-((i-x^p_k)^2 + (j - y^p_k)^2)}/{2 (\sigma_0\cdot\mathbf{s}_{k, i, j})^2}}& \\ 
		&s.t. \quad \|i - x^p_k\|_1 \leq 3\sigma \quad \|j - y^p_k\|_1 \leq 3\sigma.&
	\end{aligned}
\end{equation}
We denote the modified heatmaps as $\mathbf{H}^{\sigma_0\cdot\mathbf{s}}$. If we express $\mathbf{H}^{\sigma_0\cdot\mathbf{s}}$ by original heatmaps $\mathbf{H}^{\sigma_0}$, then we have
\begin{equation}
	\mathbf{H}^{\sigma_0\cdot\mathbf{s}}_{k, i, j}=
	\left\{
	\begin{aligned}
		& (\mathbf{H}^{\sigma_0}_{k, i, j})^{1/\mathbf{s}_{k, i, j}} \quad \mathbf{H}^{\sigma_0}_{k, i, j}>0&\\
		& \mathbf{H}^{\sigma_0}_{k, i, j}  \qquad \quad \quad \mathbf{H}^{\sigma_0}_{k, i, j}=0.&
	\end{aligned}
	\right.
\end{equation}
$\mathbf{H}^{\sigma_0\cdot\mathbf{s}}$ is what we call scale-adaptive heatmaps. It can be attained from an element-wise operation over original heatmaps, thus it is also easy to be implemented. 

For keypoints whose scale factors are larger than $1$, their corresponding standard deviation will be larger than $\sigma_0$, which means that the region covered by this Gaussian kernel will also become larger. Otherwise the reverse. Thus, to some extent, the scale factor may reflect the scale of the corresponding person.

Furthermore, some changes need to be made to stabilize the training. Firstly,  we add a regularizer loss for the predicted scale maps:
\begin{equation}
	\mathcal{L}_{regularizer} = \| (1/\mathbf{s} - 1)\mathbf{1}_{\mathbf{H}^{\sigma_0 / \mathbf{s}}>0}\|^2_2,
\end{equation}
in which $\mathbf{1}_{\mathbf{H}^{\sigma_0\cdot\mathbf{s}}>0}$ denotes the mask that keeps only regions covered by gaussian kernels. Secondly, we transform the exponential form of $\mathbf{H}^{\sigma_0\cdot\mathbf{s}}$ into a polynomial series by Taylor expansion at $\mathbf{s} =1$. We omit terms higher than the second order and have:
\begin{equation}
	\small
	\begin{aligned}
		&\qquad \qquad \mathbf{H}^{\sigma_0\cdot\mathbf{s}}_{k, i, j}=\\
		&\left\{
		\begin{aligned}
			& \frac{1}{2} \mathbf{H}^{\sigma_0}_{k, i, j}(1 + (
			1 + \boldsymbol{\alpha}_{k, i, j} \ln(\mathbf{H}^{\sigma_0}_{k, i, j}))^2) \quad \mathbf{H}^{\sigma_0}_{k, i, j}>0&\\
			&  0   \qquad \qquad \qquad \qquad \qquad \qquad \qquad \qquad \mathbf{H}^{\sigma_0}_{k, i, j}=0, &
		\end{aligned}
		\right.
	\end{aligned}
\end{equation}
where $\boldsymbol{\alpha} = 1/\mathbf{s} - 1$.
Then, the total loss is written as:
\begin{equation}
	\begin{aligned}
		\mathcal{L}_{total} &= \mathcal{L}_{regressoin}  +  \lambda \mathcal{L}_{regularizer}\\
		&= \| \mathbf{P} - \mathbf{H}^{\sigma_0\cdot\mathbf{s}}\|^2_2 + \lambda \| \boldsymbol{\alpha} \mathbf{1}_{\mathbf{H}^{\sigma_0 / \mathbf{s}}>0}\|^2_2,
	\end{aligned}
\end{equation}
where $\lambda$ is the weight for regularizer term. In practice, we use $\lambda=1$.  This is what we call scale-adaptive  heatmap regression (SAHR).

\begin{figure}
	\centering
	\includegraphics[width=\linewidth]{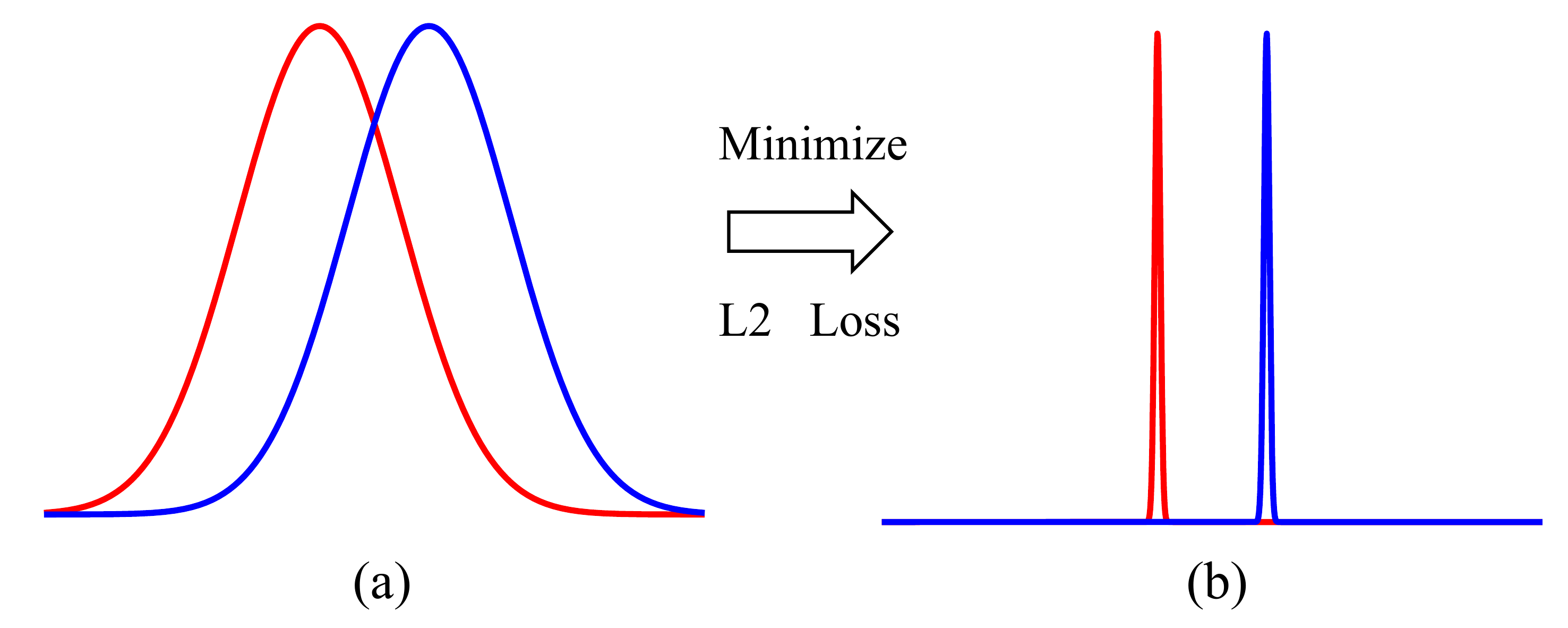}
	\caption{L2 loss cannot appropriately measure the difference between two distributions. Even if the loss is reduced a lot, the center points of these two distributions may keep unchanged. Thus original heatmap regression has to fix the standard deviation for all keypoints, and our scale-adaptive heatmap regression needs to add regularizer loss for scale map.}\label{sigma}
\end{figure}

\subsection{Relation to Uncertainty Prediction}
In~\cite{BoundingBR}, He \etal argue that there are inherent labeling ambiguities of box coordinates in some cases. Thus they treat both the predicted and ground-truth coordinates as Gaussian distributions, and the standard deviations could represent the uncertainties of the coordinates. The loss is constructed as KL loss~\cite{klloss}:
\begin{equation}
	\mathcal{L}\propto \frac{\|\mathbf{X}_p - \mathbf{X}_g \|^2_2}{2\sigma^2} + \frac{1}{2}\log(\sigma^2),
\end{equation}
where $\mathbf{X}_p$ and $\mathbf{X}_g$ denote the predicted and ground-truth coordinates respectively. And $\sigma$, which is predicted by the model, denotes the standard deviations of assumed Gaussian distributions. The former and later terms of this loss could also be treated as regression and regularizer loss respectively. It will automatically down-weight the regression loss of coordinates with relatively larger uncertainties, and thus could be more tolerant of various labeling ambiguities.

The success of the original heatmap regression could also be partially explained by the same idea. But heatmap regression uses  L2 loss instead of KL loss to measure the difference between two distributions. As shown in Figure~\ref{sigma}, simply minimizing L2 loss could not lead the model appropriately. Thus, original heatmap regression has to fix the standard deviations. However, the fixed value maybe not suitable for all keypoints and potentially hurt the performance. We still use L2 loss in SAHR. But instead of keeping the standard deviations fixed, we add a regularizer term to help lead the model to converge to the desired direction. SAHR combines the merits of both heatmap and coordinate regression.

From this perspective, the scale factor $s_{k, i, j}$ could also represent the uncertainty of the corresponding keypoint. While in the previous section we infer that scale factors indicate the scales of corresponding persons. These two statements may be consistent. The relation is also intuitive: larger persons are more likely to be labeled with larger absolute error, and thus the scale factors may be proportional to the uncertainties of corresponding persons.

\begin{table*}[t]
	\centering
	\setlength{\tabcolsep}{0.4cm}
	\resizebox{\linewidth}{!}{
		\begin{tabular}{cccccccccc}
			\hline
			\multicolumn{1}{c|}{Methods}        & \multicolumn{1}{c|}{Backbone}   & \multicolumn{1}{c|}{Input Size} & \multicolumn{1}{c|}{\#Params} & \multicolumn{1}{c|}{GFLOPs}  & AP     & $AP^{50}$ & $AP^{75}$ & $AP^{M}$ & $AP^{L}$ \\ \hline
			\multicolumn{10}{c}{w/o mutli-scale test}                                                                                                                                                                                    \\ \hline
			\multicolumn{1}{c|}{OpenPose~\cite{openpose}}      & \multicolumn{1}{c|}{-}          & \multicolumn{1}{c|}{-}          & \multicolumn{1}{c|}{-}        & \multicolumn{1}{c|}{-}       & $61.8$ & $84.9$    & $67.5$    & $57.1$   & $68.2$   \\
			\multicolumn{1}{c|}{Hourglass~\cite{stacked}}     & \multicolumn{1}{c|}{Hourglass}  & \multicolumn{1}{c|}{$512$}      & \multicolumn{1}{c|}{$277.8$}  & \multicolumn{1}{c|}{$206.9$} & $56.6$ & $81.8$    & $61.8$    & $49.8$   & $67$     \\
			\multicolumn{1}{c|}{PersonLab~\cite{personlab}}     & \multicolumn{1}{c|}{ResNet-152} & \multicolumn{1}{c|}{$1401$}     & \multicolumn{1}{c|}{$68.7$}   & \multicolumn{1}{c|}{$405.5$} & $66.5$ & $88.0$      & $72.6$    & $62.4$   & $72.3$   \\
			\multicolumn{1}{c|}{PifPaf~\cite{pipaf}}         & \multicolumn{1}{c|}{-}          & \multicolumn{1}{c|}{-}          & \multicolumn{1}{c|}{-}        & \multicolumn{1}{c|}{-}       & $66.7$ &           &           & $62.4$   & $72.9$   \\
			\multicolumn{1}{c|}{HrHRNet~\cite{higherhr}}       & \multicolumn{1}{c|}{HRNet-W32}  & \multicolumn{1}{c|}{$512$}      & \multicolumn{1}{c|}{$28.5$}   & \multicolumn{1}{c|}{$47.9$}  & $66.4$ & $87.5$    & $72.8$    & $61.2$   & $74.2$   \\
			\multicolumn{1}{c|}{HrHRNet~\cite{higherhr} + SWAHR} & \multicolumn{1}{c|}{HRNet-W32}  & \multicolumn{1}{c|}{$512$}      & \multicolumn{1}{c|}{$28.6$}   & \multicolumn{1}{c|}{$48.0$}  & $67.9$ & $88.9$    & $74.5$    & $62.4$   & $75.5$   \\
			\multicolumn{1}{c|}{HrHRNet~\cite{higherhr}}       & \multicolumn{1}{c|}{HRNet-W48}  & \multicolumn{1}{c|}{$640$}      & \multicolumn{1}{c|}{$63.8$}   & \multicolumn{1}{c|}{$154.3$} & $68.4$ & $88.2$    & $75.1$    & $64.4$   & $74.2$   \\
			\multicolumn{1}{c|}{HrHRNet~\cite{higherhr} + SWAHR} & \multicolumn{1}{c|}{HRNet-W48}  & \multicolumn{1}{c|}{$640$}      & \multicolumn{1}{c|}{$63.8$}   & \multicolumn{1}{c|}{$154.6$} & $\mathbf{70.2}$ & $\mathbf{89.9}$    & $\mathbf{76.9}$    & $\mathbf{65.2}$   & $\mathbf{77.0}$   \\ \hline
			\multicolumn{10}{c}{w/ mutli-scale test}                                                                                                                                                                                     \\ \hline
			\multicolumn{1}{c|}{Hourglass~\cite{stacked}}     & \multicolumn{1}{c|}{-}          & \multicolumn{1}{c|}{$512$}      & \multicolumn{1}{c|}{$277.8$}  & \multicolumn{1}{c|}{$206.9$} & $63.0$ & $85.7$    & $68.9$    & $58.0$   & $70.4$   \\
			\multicolumn{1}{c|}{PersonLab~\cite{personlab}}     & \multicolumn{1}{c|}{-}          & \multicolumn{1}{c|}{$1401$}     & \multicolumn{1}{c|}{$68.7$}   & \multicolumn{1}{c|}{$405.5$} & $65.5$ & $86.8$    & $72.3$    & $60.6$   & $72.6$   \\
			\multicolumn{1}{c|}{HrHRNet~\cite{higherhr}}       & \multicolumn{1}{c|}{HRNet-W48}  & \multicolumn{1}{c|}{$640$}      & \multicolumn{1}{c|}{$63.8$}   & \multicolumn{1}{c|}{$154.3$} & $70.5$ & $89.3$    & $77.2$    & $66.6$   & $75.8$   \\
			\multicolumn{1}{c|}{HrHRNet~\cite{higherhr} + SWAHR} & \multicolumn{1}{c|}{HRNet-W48}  & \multicolumn{1}{c|}{$640$}      & \multicolumn{1}{c|}{$63.8$}   & \multicolumn{1}{c|}{$154.6$} & $\mathbf{72.0}$ & $\mathbf{90.7}$    & $\mathbf{78.8}$    & $\mathbf{67.8}$   & $\mathbf{77.7}$   \\ \hline
	\end{tabular}}
	\vspace{0.0025\linewidth}
	\caption{Results on COCO test-dev2017. Top: without multi-scale test. Bottom: with multi-sale test (scale factors are $0.5$, $1.0$, and $1.5$).}\label{test-dev}
\end{table*}

\subsection{Weight-Adaptive Heatmap Regression}
We experimentally find that SAHR may aggravate the imbalance between fore-background samples in heatmap regression. This imbalance may restrict the improvement of SAHR. Most values in $\mathbf{H}^{\sigma_0\cdot\mathbf{s}}$ are zero, which may lead the model to overfit on background samples. In~\cite{focalloss}, Lin \etal propose focal loss to alleviate a similar problem in classification. It could adaptively down-weight the loss of well-classified samples and thus help the model to focus on relatively harder samples. 

To apply similar idea in heatmap regression, the straightforward way is defining a weight tensor $\mathbf{W}$ for original L2 loss:
\begin{equation}
	\mathcal{L}_{regression} = \mathbf{W} \cdot  \|\mathbf{P} - \mathbf{H} \|^2_2,
\end{equation}
And $\mathbf{W}$ can be defined as
\begin{equation}
	\small
	\mathbf{W}_{k, i, j} = \left \{
	\begin{aligned}
		&  (1 -  \mathbf{P}_{k, i, j}) \quad \{k, i, j\} \text{ is positive sample} \\
		& \mathbf{P}_{k, i, j} \qquad  \qquad  \{k, i, j\} \text{ is negative sample}
	\end{aligned}
	\right.
\end{equation}
However, in heatmap regression, the pixel values are contiguous, instead of discrete $1$ or $0$, thus it is difficult to determine which are positive (negative) samples.

Towards this issue, we propose a weight-adaptive heatmap regression (WAHR), in which the loss weights are written as:
\begin{equation}
	\mathbf{W} =(\mathbf{H})^{\gamma} \cdot \| 1 - \mathbf{P} \|+ \| \mathbf{P} \|\cdot (1 - (\mathbf{H})^{\gamma})\\
\end{equation}
where $\gamma$ is the hyper-parameter that controls the position of a \textit{soft boundary}. And the soft boundary is defined as a threshold heatmap value $p$, where $1 - p^{\gamma} = p^{\gamma}$. For samples with heatmap values larger than $p$, their loss weights are more close to ($1 -\mathbf{P}$), otherwise are more close to $\mathbf{P}$. We can get the threshold $p=2^{-\frac{1}{\gamma}}$. In practice, we use $\gamma=0.01$.

Experiments in Sec~\ref{aba_section} show that WAHR can further advance the improvement of  SAHR. When SAHR and WAHR are used together, we call it the scale and weight adaptive heatmap regression (SWAHR). 

\begin{table}[t]
	\centering
	\setlength{\tabcolsep}{0.15cm}
	\resizebox{\linewidth}{!}{
		\begin{tabular}{cccccc}
			\hline
			\multicolumn{1}{c|}{Methods}             & $AP$   & $AP^{50}$ & $AP^{75}$ & $AP^{M}$ & $AP^{L}$ \\ \hline
			\multicolumn{6}{c}{Top-down methods}                                                           \\ \hline
			\multicolumn{1}{c|}{Mask-RCNN~\cite{maskrcnn}}          & $63.1$ & $87.3$    & $68.7$    & $57.8$   & $71.4$   \\
			\multicolumn{1}{c|}{G-RMI~\cite{grmi}}              & $64.9$ & $85.5$    & $71.3$    & $62.3$   & $70.0$   \\
			\multicolumn{1}{c|}{Sun \etal~\cite{integralpose}}           & $67.8$ & $88.2$    & $74.8$    & $63.9$   & $74.0$   \\
			\multicolumn{1}{c|}{G-RMI~\cite{grmi} + extra data} & $68.5$ & $87.1$    & $75.5$    & $65.8$   & $73.3$   \\
			\multicolumn{1}{c|}{CPN~\cite{cpn}}                & $72.1$ & $91.4$    & $80.0$    & $68.7$   & $77.2$   \\
			\multicolumn{1}{c|}{RMEPE~\cite{rmepe}}              & $72.3$ & $89.2$    & $79.1$    & $68.0$   & $78.6$   \\
			\multicolumn{1}{c|}{CFN~\cite{cfn}}                & $72.6$ & $86.1$    & $69.7$    & $78.3$   & $64.1$   \\
			\multicolumn{1}{c|}{CPN(ensemble)~\cite{cpn}}      & $73.0$ & $91.7$    & $80.9$    & $69.5$   & $78.1$   \\
			\multicolumn{1}{c|}{SimpleBaseline~\cite{simplebaseline}}     & $73.7$ & $91.9$    & $81.1$    & $70.3$   & $80.0$   \\
			\multicolumn{1}{c|}{HRNet-W48~\cite{hrnet}}          & $75.5$ & $92.5$    & $83.3$    & $71.9$   & $81.5$   \\ \hline
			\multicolumn{6}{c}{Bottom-up methods}                                                          \\ \hline
			\multicolumn{1}{c|}{OpenPose~\cite{openpose}}           & $61.8$ & $84.9$    & $67.5$    & $57.1$   & $68.2$   \\
			\multicolumn{1}{c|}{Hourglass~\cite{stacked}}          & $65.5$ & $86.8$    & $72.3$    & $60.6$   & $72.6$   \\
			\multicolumn{1}{c|}{PifPaf~\cite{pipaf}}             & $66.7$ &    -  &      -     & $62.4$   & $72.9$   \\
			\multicolumn{1}{c|}{SPM~\cite{spm}}                & $66.9$ & $88.5$    & $72.9$    & $62.6$   & $73.1$   \\
			\multicolumn{1}{c|}{PersonLab~\cite{personlab}}          & $68.7$ & $89.0$    & $75.4$    & $64.1$   & $75.5$   \\
			\multicolumn{1}{c|}{HrHRNet-W48~\cite{higherhr}}       & $70.5$ & $89.3$    & $77.2$    & $66.6$   & $75.8$   \\
			\multicolumn{1}{c|}{HrHRNet-W48~\cite{higherhr} + SWAHR}  & $\mathbf{72.0}$ & $\mathbf{90.7}$    & $\mathbf{78.8}$    & $\mathbf{67.8}$   & $\mathbf{77.7}$   \\ \hline
	\end{tabular}}
	\vspace{0.005\linewidth}
	\caption{Results on COCO test-dev2017. Top: top-down methods. Bottom: bottom-up methods (with multi-scale test).} \label{top-down}
\end{table}

\subsection{Implementation Details}
\begin{figure}[t]
	\centering
	\includegraphics[width=\linewidth]{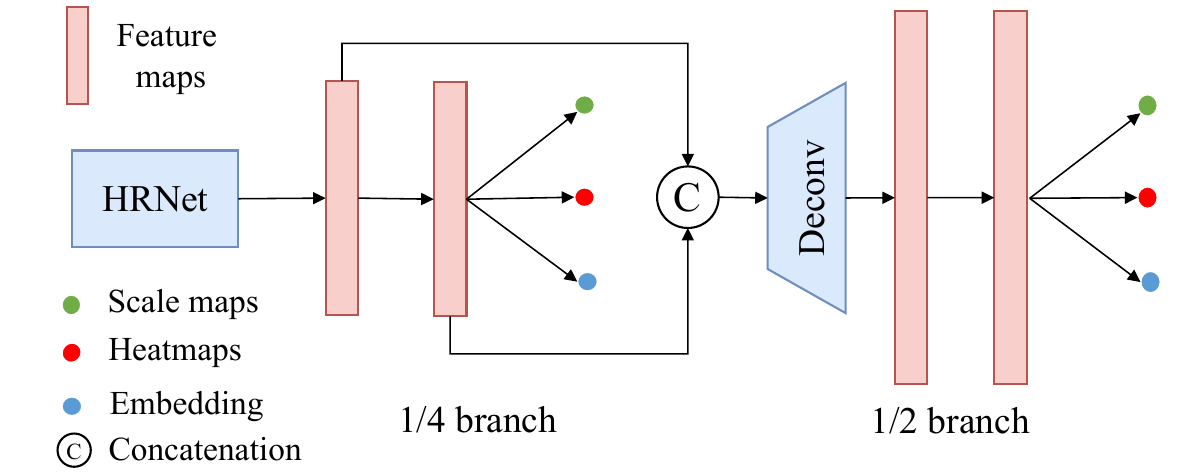}
	\caption{Implementation of scale and weight adaptive heatmap aggregation on HrHRNet.}\label{hrhrnet}
\end{figure}

In this paper, we mainly implement the proposed heatmap regression on HrHRNet~\cite{higherhr}, which is a HRNet~\cite{hrnet} with deconvolution modules. As shown in Figure~\ref{hrhrnet}, it predicts multi-scale heatmaps, which are $1/4$ and $1/2$ sizes of the original image respectively. During training, these two branches are independently supervised by different heatmaps. During testing, it aggregates multi-scale heatmaps to form the final predictions. The larger size of heatmaps largely benefits the accuracy of keypoints detection, and the heatmaps aggregation helps the model achieve remarkable results with only a single-scale test. The grouping is done by associate embedding~\cite{ae}. For SAHR we add an extra branch to predict scale maps, and the model is denoted as HrHRNet + SAHR. If only WAHR is used, the model is denoted as HrHRNet + WAHR. And if both methods are used, the model is denoted as HrHRNet  + SWAHR. 

\section{Experiments}
\subsection{COCO Keypoint Detection}
\textbf{Dataset.} Most of our experiments are done COCO dataset~\cite{coco}, which contains over $200$K images and $250$K person instances. Our models are trained on COCO train2017 ($57$K images), and evaluated on COCO val2017 ($5$K images) and COCO test-dev ($20$K images).

\textbf{Evaluation metric.} We use the standard evaluation metric Object Keypoint Similarity (OKS) to evaluate our models. $\text{OKS}={\frac{\sum_{i}\exp(-d_i^2/2s^2k_i^2)\delta(v_i>0)}{\sum_{i}\delta(v_i>0)}}$, where $d_i$ is the Euclidean distance between the detected keypoint and its corresponding ground-truth, $v_i$ is the visibility flag of the ground-truth, $s$ denotes the person scale, and $k_i$ is a per-keypoint constant that controls falloff. We report the standard average precision ($AP$) and recall, including $AP^{50}$ ($AP$ at OKS=$0.5$), $AP^{75}$, $AP$ (mean of $AP$ scores from OKS=$0.50$ to OKS=$0.95$ with the increment as $0.05$, $AP^{M}$ ($AP$ scores for person of medium sizes) and $AP^{L}$ ($AP$ scores for persons of large sizes).

\textbf{Training.} Following the setting of~\cite{ae,higherhr}, we augment the data by random rotation ([$-30^\circ$, $30^\circ$]), random scaling ([$0.75$, $1.25$]), random translation ([$-40$, $40$]) and random horizontal flip. The input image is then cropped to $512\times512$ (or $640\times640$). 

The models are optimized by Adam~\cite{adam} optimizer, and the initial learning rate is set as $2\times10^{-3}$. Each model is trained for $300$ epochs and the learning rate will linearly decay to $0$ in the end.

\textbf{Testing.} Following the setting of~\cite{higherhr}, the input image is firstly padded to square and then resize the short side to $512$ (or $640$). We also perform heatmap aggregation by averaging output heatmaps of different sizes. The flip test is also performed in all experiments. For the multi-scale test, we resize the original image by scale factor $0.5$, $1.0$, and $1.5$ respectively, and then aggregate the heatmaps as the final prediction.

\textbf{Results on COCO test-dev2017.} We firstly make comparisons with the state-of-the-art bottom-up HPE methods. Results are shown in Table~\ref{test-dev}. As one can see, with the help of SWAHR, HrHRNet can achieve the best results with or without multi-scale test. And if with multi-scale test, it can finally achieve $\mathbf{72.0}$ AP score on test-dev2017. On the other hand, SWAHR can bring steady improvements to HrHRNets with different backbones and different input sizes, while introducing only marginal computational cost.

Then we make comparisons with recent top-down HPE methods. Results are shown in Table~\ref{top-down}. As one can see, with the help of SWAHR, HrHRNet-W48 has exceeded many early top-down methods. CPN~\cite{cpn} is the champion of COCO Keypoint Challenge in 2017, and our method gets nearly the same results as it.

\begin{figure*}[t]
	\centering
	\includegraphics[width=\linewidth]{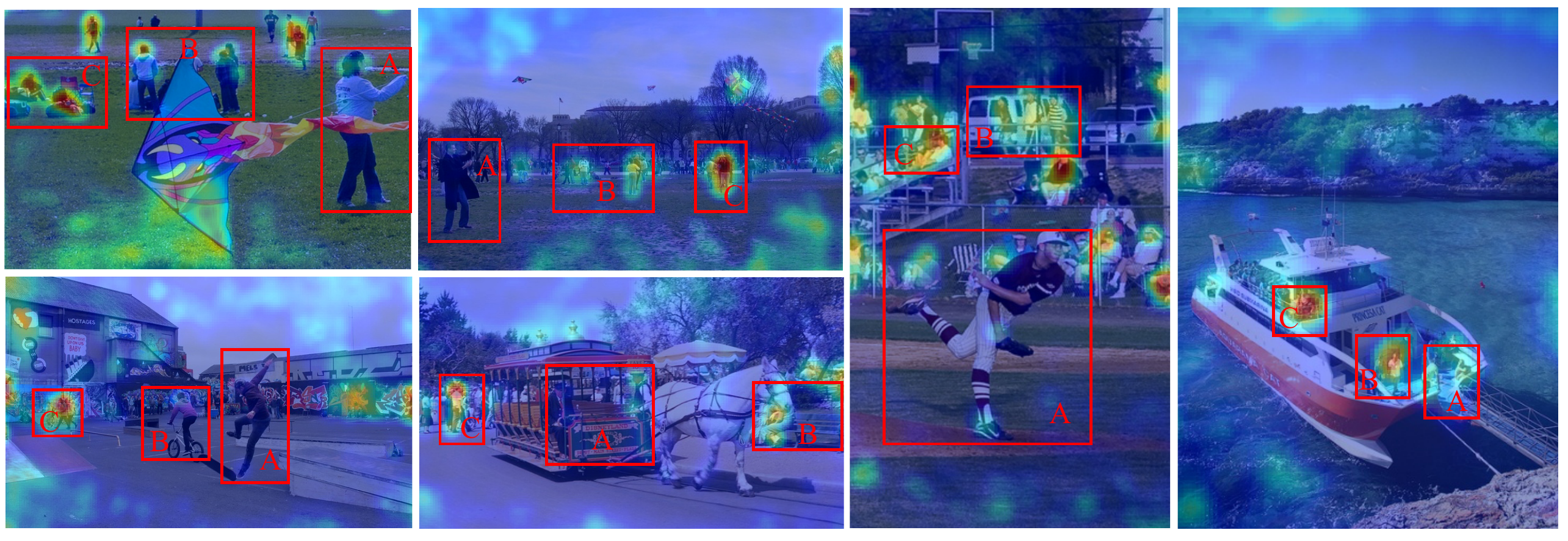}
	\caption{Visualization of $1/\mathbf{s}$. Redder pixels indicate larger values \ie smaller scale factors. Within each image, the order of person scales ($A \textgreater B \textgreater C$)  is usually the same as the order of scale factors ($A \textgreater B \textgreater C$).}\label{scale_map}
\end{figure*}

\subsection{Analysis}
\textbf{Ablation study.}\label{aba_section} We design comparative experiments to validate the improvement brought by SAHR and WAHR respectively. We use HrHRNet-W32 as the baseline model, and validate these models on COCO val2017 dataset without multi-scale test. As we can see in Table~\ref{aba_study}, SAHR can bring an improvement of $+0.7 AP$. If WAHR is further added, they together could bring an improvement of $+1.8 AP$. Also, WAHR  alone can bring improvement of $+1.3 AP$ improvements respectively. This is because the server imbalance between fore-background samples also exists in original heatmap regression.

Looking into the improvements on different scales, we can see that WAHR can largely benefit the keypoints detection of both medium and large persons. This is easy to understand: the severe imbalance between fore-background samples exists both for large and medium persons, thus WAHR could benefit both. Since the original base standard deviation is only suitable for relatively larger persons (Figure~\ref{nose_heatmap} (a)), SAHR mainly focuses on adjusting the standard deviations for relatively smaller persons. Thus, the improvements brought by SA are mainly attributed to better performance on medium persons. 

\begin{table}[h]
	\centering
	\setlength{\tabcolsep}{0.6cm}
	\resizebox{\linewidth}{!}{
		\begin{tabular}{cc|ccc}
			\hline
			SAHR      & WAHR      & $AP$   & $AP^{M}$ & $AP^{L}$ \\ \hline
			&         & $67.1$ & $61.5$   & $76.1$   \\
			& $\surd$        & $68.4$ & $62.5$   & $77.0$   \\ 
			$\surd$  &       & $67.8$ & $62.5$   & $76.1$   \\
			$\surd$ & $\surd$ & $68.9$ & $63.0$   & $77.5$   \\ \hline
	\end{tabular}}
	\vspace{0.02\linewidth}
	\caption{Ablation study on SAHR and WAHR respectively. The results are reported on COCO val2017 dataset. Single-scale test only.} \label{aba_study}
\end{table}

\textbf{Visualizing scale maps.} We visualize the learned scale maps $1/\mathbf{s}$ in Figure~\ref{scale_map}. We resize the maps to the same size as the original image, take mean values along the channel dimension,  and normalize the maps along spatial dimensions. The colormaps are drawn as that redder pixels indicate larger values, which also means smaller scale factors. As one can see, on the whole, smaller persons usually have smaller scale factors. For the scales of persons in boxes $A$, $B$, and $C$, we have $A \textgreater B \textgreater C$. While for scale factors in these boxes, we can also get the same order. It suggests that SAHR adaptively down-scales the standard deviations for keypoints of smaller persons, while up-scales that for relatively larger ones. Without extra supervision, SAHR has learned the relationship between human scales and the suitable standard deviations. 

\textbf{Naive baseline.} Although the standard deviation for each keypoints is not labeled. The scale for each person could be roughly deduced from the bounding box. As a comparison, we substitute the scale maps in SAHR with the deduced scale factor as a naive baseline. We call it scaled heatmap regression (SHR). Specifically, we manually calculate the scale factor $s_{k, i, j} = W_{box}/W_{base}$, where $W_{box}$ denotes the width of the bounding box of corresponding person, and $W_{base}$ is the base width. In practice, we use $W_{base} = 256$. We use HrHRNet-W32 as the baseline. The comparative results are shown in Table~\ref{naive_baseline}. SHR even hurts the performance of the original model. This is easy to explain: the width of bounding boxes can not exactly represent the person scales, because of various poses and occlusions. This naive implementation may cause more confusion, and thus hurt the original performance. Instead, SAHR could avoid this case by additively learning the scale maps.  

\begin{table}[h]
	\centering
	\setlength{\tabcolsep}{0.35cm}
	\resizebox{\linewidth}{!}{
		\begin{tabular}{c|ccccc}
			\hline
			Methods & $AP$ & $AP^{0.5}$ & $AP^{0.75}$ & $AP^{M}$ & $AP^{L}$ \\
			\hline
			baseline & $67.1$ & $86.2$& $73.0$ &$61.5$ & $76.1$  \\
			SHR & $63.9$ & $84.7$& $68.0$ &$55.5$ & $77.0$  \\
			SAHR  & $67.8$ & $86.8$& $73.7$ &$62.5$ & $76.1$  \\
			\hline
	\end{tabular}}
	\vspace{0.02\linewidth}
	\caption{Comparison with naive baseline on COCO val2017 without multi-scale test.} \label{naive_baseline}
\end{table}

\textbf{Study of $\mathbf{\lambda}$.} In SAHR, there is a hyper-parameter $\lambda$, \ie the weight for regularizer loss. Larger $\lambda$ means that the model has to `pay more' to adjust the standard deviation. It indicates that we are more confident about the manually constructed ground-truth heatmaps. And smaller $\lambda$ indicates that we encourage the model to adjust the standard deviation for each keypoint by itself. We compare the performance of HrHRNet-W32 + SAHR with different $\lambda$ on COCO val2017 dataset. As shown in Table~\ref{lambda}, when $\lambda=+\infty$, which means that the model is not allowed to adjust the standard deviations, it will degrade to the original baseline model. On the whole, the improvement brought by SA is no too sensitive to $\lambda$, as the results keep the same when $\lambda=1$ and $\lambda=0.5$. But when $\lambda$ becomes too small, \ie $\lambda=0.1$, the model may be able to largely adjust the standard deviations, while the model may be not reliable enough. In such cases, the improvements may get hurt. 

\begin{table}[h]
	\centering
	\setlength{\tabcolsep}{0.4cm}
	\resizebox{0.8\linewidth}{!}{
		\begin{tabular}{c|cccc}
			\hline
			$\lambda$ & $0.1$  & $0.5$  & $1.0$ &$+\infty$  \\ \hline
			$AP$     & $67.6$ & $67.8$ & $67.8$ &$67.1$ \\ \hline
	\end{tabular}}
	\vspace{0.02\linewidth}
	\caption{Study of hyper-parameter $\lambda$. Results are reported on COCO val2017 dataset, without WA and multi-scale test.}\label{lambda}
\end{table}

\textbf{Study of $\mathbf{\gamma}$.} In WAHR, the hyper-parameter $\gamma$ controls the soft boundary between positive and negative samples. Smaller $\gamma$ indicates that more samples will be determined as positive ones. To investigate the influence of $\gamma$, we compare the performance of HrHRNet-W32 + WAHR with different $\gamma$ on COCO val2017 dataset. As shown in Table~\ref{gamma}, when $\gamma$ decreases, the $AP$ score firstly grows quickly, and then get stable at $68.4$, when $\gamma=0.001$. As $\gamma$ decreases, the threshold value $p$ will also exponentially decrease. When $\gamma=0.01$, $p\approx8\times10^{-31}$. In that case, almost all regions that are covered by gaussian kernels have heatmap values larger than $p$. Thus, a further decrement of $\gamma$ makes little difference to the final results.

\begin{table}[h]
	\centering
	\setlength{\tabcolsep}{0.4cm}
	\resizebox{0.8\linewidth}{!}{
		\begin{tabular}{c|cccc}
			\hline
			$\gamma$ & $1.0$ & $0.1$  & $0.01$  & $0.001$  \\ \hline
			$AP$     & $67.8$ & $68.2$ & $68.4$ & $68.4$ \\ \hline
	\end{tabular}}
	\vspace{0.02\linewidth}
	\caption{Study of hyper-parameter $\gamma$. Results are reported on COCO val2017 dataset, without SA and multi-scale test.}\label{gamma}
\end{table}

\textbf{Larger receptive filed V.S.  Larger $\mathbf{\sigma}$.} The intuitive idea is that a larger receptive field will benefit the accuracy of larger persons. In this section,  we experimentally illustrate that the accuracy of larger persons may be more related to larger standard deviations that are used to construct ground-truth heatmaps. 

We first compare the results with different receptive fields. We still use HrHRNet-W32 as the baseline model. To exclude the influence of heatmaps aggregation, we only use the results of larger heatmaps ($1/2$ size of the original image). There are $4$ residual blocks in this branch. We change the dilation rates of their convolutional layers to change the sizes of their receptive fields. Different models are denoted as $dddd$, where each $d$ denotes the dilation rates of the corresponding residual block. The baseline model is denoted as $1111$. Then we change it to $1122$ and $2222$ to investigate the influence. As shown in Table~\ref{dilation}, as the dilation rates increases, the AP scores of large persons almost keep the same. It indicates that the accuracies of large persons are not restricted by the sizes of receptive fields.

Then we investigate the influence of standard deviations that are used to construct the ground-truth heatmaps. Results of the $1/2$ branch of HrHRNet-W32 are reported in Table~\ref{table_sigma}. As one can see, with an increase of  $\sigma$, the performance on medium persons becomes worse, while the model performs better on large persons. It suggests that a larger $\sigma$ is more suitable for larger persons. This is also consistent with our previous assumption: keypoints of larger persons have larger semantically discriminative regions and also larger labeling ambiguities.

\begin{table}[h]
	\centering
	\setlength{\tabcolsep}{0.4cm}
	\resizebox{0.6\linewidth}{!}{
		\begin{tabular}{c|cccc}
			\hline
			dilation & $1111$ & $1122$  & $2222$   \\ \hline
			$AP$     & $66.6$ & $66.6$ & $66.7$\\ \hline
			$AP^{M}$     & $61.3$ & $61.4$ & $61.3$\\ \hline
			$AP^{L}$     & $75.0$ & $75.0$ & $75.1$ \\ \hline
	\end{tabular}}
	\vspace{0.02\linewidth}
	\caption{Study of receptive fields. Results are reported on COCO val2017 dataset, without SA and multi-scale test.}\label{dilation}
\end{table}

\begin{table}[h]
	\centering
	\setlength{\tabcolsep}{0.4cm}
	\resizebox{0.6\linewidth}{!}{
		\begin{tabular}{c|cccc}
			\hline
			$\sigma$ & $2$ & $2.5$  & $3$   \\ \hline
			$AP$     & $66.6$ & $66.1$ & $65.4$\\ \hline
			$AP^{M}$     & $61.3$ & $60.1$ & $58.3$\\ \hline
			$AP^{L}$     & $75.0$ & $75.2$ & $75.4$ \\ \hline
	\end{tabular}}
	\vspace{0.02\linewidth}
	\caption{Study of receptive fields. Results are reported on COCO val2017 dataset, without SA and multi-scale test.}\label{table_sigma}
\end{table}


\subsection{CrowdPose}

We further make comparisons with state-of-the-art HPE methods on CrowdPose dataset~\cite{crowdpose}. It contains about 20000 images and 80000 person instances The training, validation, and testing datasets contain about 10000, 2000, and 8000 images respectively. CrowdPose dataset has more crow cases than COCO~\cite{coco}, and thus is more challenging to multi-person pose estimation. The evaluation metric almost the same as that of COCO, but with extra AP scores on relatively easier samples ($AP^E$) and relatively harder samples ($AP^H$).

We firstly make comparisons with top-down methods. As shown in Table~\ref{crowdpose}, top-down methods have lost their superiority in crowd scenes. This is because top-down methods assume that all persons could be completely copped by the human detector, and each crop contains only one person. However, this assumption does not hold in crowd scenes, where persons are usually heavily overlapped. While bottom-up methods do not rely on the human detector and may be better at tackling crowd scenes.

Based on HrHRNet, SWAHR could bring $+5.7AP$ improvements without multi-scale test, and $+6.2AP$ with multi-scale test. which are much more significant on COCO test-dev (Table~\ref{test-dev}). It indicates that SWAHR could bring more improvements in crowd scenes. This may because that SWAHR has taken the various human scales into considerations, and this problem is more evident in crow scenes.

\begin{table}[h]
	\centering
	\setlength{\tabcolsep}{0.15cm}
	\resizebox{\linewidth}{!}{
		\begin{tabular}{ccccccc}
			\hline
			\multicolumn{1}{c|}{Methods}                &      $AP$       &    $AP^{50}$    &    $AP^{75}$    &    $AP^{E}$     &    $AP^{M}$     &    $AP^{H}$     \\ \hline
			\multicolumn{7}{c}{Top-down methods}                                                                  \\ \hline
			\multicolumn{1}{c|}{Mask-RCNN~\cite{maskrcnn}}       &     $57.2$      &     $83.5$      &     $60.3$      &     $69.4$      &     $57.9$      &     $45.8$      \\
			\multicolumn{1}{c|}{AlphaPose~\cite{rmepe}}        &     $61.0$      &     $81.3$      &     $66.0$      &     $71.2$      &     $61.4$      &     $51.1$      \\
			\multicolumn{1}{c|}{SimpleBaseline~\cite{simplebaseline}} &     $60.8$      &     $84.2$      &     $71.5$      &     $71.4$      &     $61.2$      &     $51.2$      \\ \hline
			\multicolumn{7}{c}{Top-down with refinement}                                                              \\ \hline
			\multicolumn{1}{c|}{SPPE~\cite{crowdpose}}         &     $66.0$      &     $84.2$      &     $71.5$      &     $75.5$      &     $66.3$      &     $57.4$      \\ \hline
			\multicolumn{7}{c}{Bottom-up methods w/o multi-scale testt}                                                      \\ \hline
			\multicolumn{1}{c|}{OpenPose~\cite{openpose}}       &        -        &        -        &        -        &     $62.7$      &     $48.7$      &     $32.3$      \\
			\multicolumn{1}{c|}{HrHRNet-W48~\cite{higherhr}}      &     $65.9$      &     $86.4$      &     $70.6$      &     $73.3$      &     $66.5$      &     $57.9$      \\
			\multicolumn{1}{c|}{HrHRNet-W48~\cite{higherhr} + SWAHR}  & $\mathbf{71.6}$ & $\mathbf{88.5}$ & $\mathbf{77.6}$ & $\mathbf{78.9}$ & $\mathbf{72.4}$ & $\mathbf{63.0}$ \\ \hline
			\multicolumn{7}{c}{Bottom-up methods w/ multi-scale testt}                                                       \\ \hline
			\multicolumn{1}{c|}{HrHRNet-W48~\cite{higherhr}}      &     $67.6$      &     $87.4$      &     $72.6$      &     $75.8$      &     $68.1$      &     $58.9$      \\
			\multicolumn{1}{c|}{HrHRNet-W48~\cite{higherhr} + SWAHR}  & $\mathbf{73.8}$ & $\mathbf{90.5}$ & $\mathbf{79.9}$ & $\mathbf{81.2}$ & $\mathbf{74.7}$ & $\mathbf{64.7}$ \\ \hline
	\end{tabular}}
	\vspace{0.002\linewidth}
	\caption{Comparisons with top-down and bottom up methods on CrowPose test dataset.}\label{crowdpose}
	\vspace{-0.01\linewidth}
\end{table}

\section{Conclusion}
In this paper, we mainly focus on the problems in heatmap regression when tackling various human scales and labeling ambiguities. We argue that in the ground-truth heatmaps, keypoints of relatively larger persons should be covered by gaussian kernels with also relatively larger standard deviation. We illustrate this problem from the perspectives both of semantically discriminative regions labeling ambiguities. Towards this issue, we propose a scale-adaptive heatmap regression (SAHR), which can learn to adjust the standard deviation for each keypoint by itself. Without extra supervision, experiments show that the model could learn the relation between standard deviation and the corresponding human scales. Also, as SAHR may aggravate the imbalance between fore-background samples, we propose a weight-adaptive heatmap regression (WAHR) to alleviate this problem. WAHR could automatically down-weight the loss of well-classified samples and focus more on relatively harder  (usually foreground) samples. Experiments show that the two methods (SAHR and WAHR) together can largely improve the performance of the original model. As a result, we finally outperform the state-of-the-art model by  $+1.5 AP$ and achieve $72.0 AP$ on COCO test-dev2017 dataset, which is comparable with the performances of most top-down methods.

\section*{Acknowledgements}
This paper is supported by the National Key R\&D Plan of the Ministry of Science and Technology (“Grid function expansion technology and equipment for community risk prevention”, Project No. 2018YFC0809704).

{\small
	\bibliographystyle{ieee_fullname}
	
}

\end{document}